\pdfoutput=1
\documentclass[11pt]{article}

\usepackage[final]{acl}

\usepackage{times}
\usepackage{latexsym}
\usepackage{amsfonts}
\usepackage{amsmath}
\usepackage{subcaption}
\usepackage{graphicx}
\usepackage{scalefnt} %用于设置字体

\usepackage{graphicx}
\usepackage{xcolor}
\usepackage{CJKutf8}
\usepackage{tabularx}
\usepackage{booktabs}
\usepackage{makecell}
\usepackage[misc]{ifsym}
\usepackage{amssymb}% http://ctan.org/pkg/amssymb
\usepackage{pifont}% http://ctan.org/pkg/pifont
\usepackage[export]{adjustbox}
\usepackage{multirow}

\newcommand{\cmark}{\ding{51}}%
\newcommand{\xmark}{\ding{55}}%

\usepackage[T1]{fontenc}
\usepackage[utf8]{inputenc}

\usepackage{microtype}

\usepackage{inconsolata}

\usepackage{graphicx}

\title{MiLiC-Eval: Benchmarking Multilingual LLMs\\for China's Minority Languages}

\author{Chen Zhang,\ \ Mingxu Tao,\ \  Zhiyuan Liao,\ \ Yansong Feng\thanks{Corresponding author.} \\
Wangxuan Institute of Computer Technology, Peking University \\
{\tt \{zhangch,thomastao,fengyansong\}@pku.edu.cn} \\
{\tt liaozy@stu.pku.edu.cn}\\
}

\begin{document}
\maketitle
\begin{abstract}
Large language models (LLMs) excel in high-resource languages but struggle with low-resource languages (LRLs), particularly those spoken by minority communities in China, such as Tibetan, Uyghur, Kazakh, and Mongolian. To systematically track the progress in these languages, we introduce MiLiC-Eval, a benchmark designed for minority languages in China, featuring 24K instances across 9 tasks. MiLiC-Eval focuses on underrepresented writing systems. Its parallelism between tasks and languages can provide a faithful and fine-grained assessment of linguistic and problem-solving skills. Our evaluation reveals that open-source LLMs perform poorly on syntax-intensive tasks and multi-script languages. We further demonstrate how MiLiC-Eval can help advance LRL research in handling diverse writing systems and understanding the process of language adaptation~\footnote{Our data and code are available at \url{https://github.com/luciusssss/MiLiC-Eval}.}.
\end{abstract}

\section{Introduction}

Large language models (LLMs) have achieved remarkable success in high-resource languages such as English and Chinese, demonstrating the ability to perform sophisticated tasks, including creative writing~\cite{gomez-rodriguez-williams-2023-confederacy}, complex planning~\cite{huang2024understanding}, and scientific reasoning~\cite{wang2024scibench,jaech2024openai,guo2025deepseek}.
However, thousands of low-resource languages (LRLs) remain under-explored by LLMs, which exhibit significant limitations in their basic linguistic capabilities for these languages~\cite{alam-etal-2024-llms}.

This challenge is particularly pronounced for minority languages spoken in China, such as Tibetan (bo), Uyghur (ug), Kazakh (kk), and Mongolian (mn). Although spoken by tens of millions, these languages are largely marginalized in NLP research due to their limited digital representation and the scarcity of training data~\cite{zhang-etal-2024-mc2}. 
Furthermore, their use of non-Latin scripts, including non-mainstream ones like those in Mongolian and Kazakh, poses additional challenges for tokenization and language modeling.

\begin{figure}[t]
\centering
  \includegraphics[width=\columnwidth]{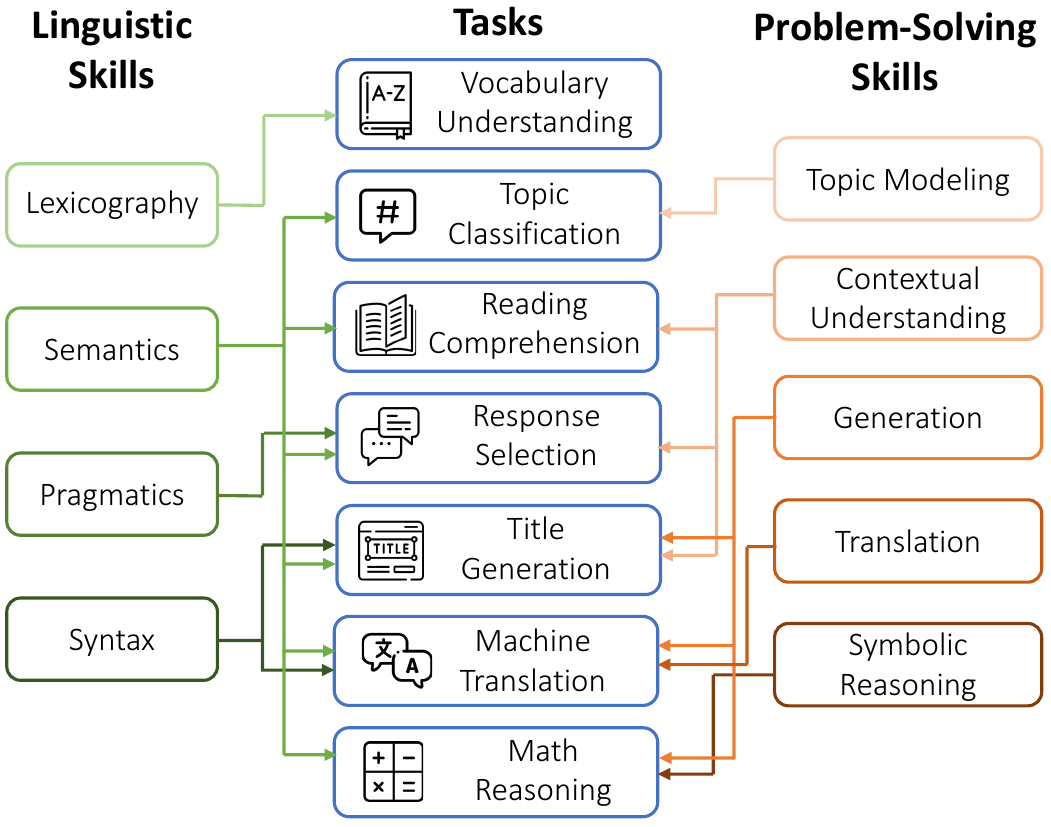}
  \caption{The linguistic and problem-solving skills that the tasks in MiLiC-Eval assess.}
  \label{fig:task_skills}
\end{figure}

Currently, there are no standardized benchmarks for evaluating LLMs on these languages. 
To fill this gap, we introduce \textbf{MiLiC-Eval}, a comprehensive framework with 24K instances across 9 tasks, focusing on 4 \underline{Mi}nority \underline{L}anguages \underline{i}n \underline{C}hina. 
It overcomes limitations in previous LRL benchmarks such as inadequate attention to low-resource writing systems and unsystematic task lineups~\cite{hu2020xtreme,ahuja-etal-2023-mega,zhang2024p}. MiLiC-Eval offers the following key features:
(1) Focus on underrepresented writing systems such as traditional Mongolian and Tibetan scripts;
(2) Faithful evaluation with parallel tasks across languages and formats;
(3) Fine-grained skill evaluation, assessing both linguistic and problem-solving abilities, from vocabulary to symbolic reasoning, as shown in Figure~\ref{fig:task_skills}.

Our evaluation of open-source multilingual LLMs reveals that they especially struggle with languages using less common writing systems, such as Mongolian. 
Skill-wise evaluation shows that LLMs have basic lexical knowledge in these languages and can perform simple tasks such as topic modeling. 
However, they still lack the abilities for generation and translation, which require intensive syntactic knowledge.

Finally, we discuss how MiLiC-Eval can serve as a valuable resource for advancing LRL research. 
First, MiLiC-Eval can facilitate the development of multilingual LLMs that are inclusive and robust across diverse writing systems. 
Our findings highlight inefficiencies in current LLMs, particularly in tokenization and the handling of multiple scripts, which pose significant challenges for LRLs with unique or complex writing systems.
Second, MiLiC-Eval provides a faithful and reliable way to evaluate models' abilities in LRLs. Its task-parallelism design prevents biased conclusions resulting from over-reliance on a single task format. The human-translated data avoids the misinterpretation of model performance caused by the noise in machine-translated data.
Third, MiLiC-Eval enables a deeper investigation into how LLMs acquire and exhibit various abilities during language adaptation. Using MiLiC-Eval, we reveal limitations in prevailing practices, calling for more effective techniques for LRL adaptation.

Our main contributions are as follows:
(1) We introduce MiLiC-Eval, a standardized benchmark for LRLs in China, emphasizing underrepresented writing systems.
(2) We provide a skill-wise evaluation of state-of-the-art multilingual LLMs, uncovering their limitations in handling syntax-intensive tasks and multiple writing systems.
(3) We demonstrate the utility of MiLiC-Eval in advancing LRL research, offering transferable insights for the study of LRLs in other underrepresented regions.

\begin{table*}[t]
\small
\centering
\setlength\tabcolsep{4.5pt}
\begin{tabular}{l|cc|c|cc|c}
\toprule
\textbf{Task}  & \textbf{\# Per Lang.} &  \textbf{Train / Dev / Test}  & \textbf{Domain} & \textbf{Task Para.}& \textbf{Lang. Para.} & \textbf{Metric}\\
\midrule
Vocabulary Understanding & 1,000 & 20 * 3 / 40 / 900 & General &  & \xmark & Accuracy \\
Topic Classification (Sentence) & \ \ 492 & 10 * 3 / 30 / 432 &  Wiki & $\clubsuit$ & \cmark & Accuracy \\
Topic Classification (Passage) & \ \ 600 & 16 * 3 / 48 / 504 & News &  & \xmark & Accuracy \\
Reading Comprehension & \ \ 250 & 10 * 3 / 20 / 200& Dialogue & $\spadesuit$ & \cmark & Accuracy \\
Response Selection  & \ \ 507 & 20 * 3 / 40 / 407 & Dialogue & $\spadesuit$ & \cmark & Accuracy\\
Title Generation  &  1,000 & 20 * 3 / 40 / 900 & News & & \xmark & ROUGE-L \\
Machine Translation (Article)  &  1,012 & 20 * 3 / 40 / 912 & Wiki & $\clubsuit$ & \cmark & chrF++\\ 
Machine Translation (Dialogue)  &  \ \ 773 & 20 * 3 / 40 / 673 & Dialogue & $\spadesuit$ & \cmark & chrF++\\
Math Reasoning  &  \ \  250 & 10 * 3 / 20 / 200 & Textbook & & \cmark & Accuracy\\
\bottomrule
\end{tabular}
\caption{Statistics, characteristics, and metrics of each task in MiLiC-Eval. The tasks sharing the same symbols in \textbf{Task Para.} are constructed from the same sets of texts.
\textbf{Lang. Para.} denotes whether the data are parallel across 6 languages, including Chinese, English, and the four minority languages. }
\label{tab:dataset_statistics}
\end{table*}

\section{MiLiC-Eval}

We introduce MiLiC-Eval, the first standardized benchmark for evaluating LLMs' performance on four minority languages in China. 
The benchmark includes 9 distinct tasks with 24K instances.
Basic information about each task is provided in 
Table~\ref{tab:dataset_statistics}.

\subsection{Design Principles}

\paragraph{Underrepresented Writing Systems}
While current LLMs have made encouraging progress in supporting LRLs~\cite{ji2024emma500,yang2025qwen3}, there has been limited focus on improving their handling of diverse, less common writing systems, particularly those that vary across communities.

The four languages in MiLiC-Eval all use non-Latin scripts, which are poorly supported by English-dominant LLMs. 
Notably, MiLiC-Eval includes two languages that adopt different writing systems in different communities: The Kazakh and Mongolian communities in China use the Arabic Kazakh script and traditional Mongolian script, respectively, instead of the \textit{mainstream} Cyrillic script.
We are the first to benchmark these less-common writing systems.
MiLiC-Eval can serve as a testbed for evaluating LLMs' ability to handle less-represented writing systems.

\paragraph{Cross-Language and Cross-Task Parallelism}
Many existing multi-task benchmarks consist of separate datasets covering different languages~\cite{asai-etal-2024-buffet}, often leading to underrepresentation of LRLs. Besides, the tasks in these benchmarks are typically simple natural language understanding tasks~\cite{hu2020xtreme}. 

MiLiC-Eval is highly parallel across both languages and tasks. For six tasks in MiLiC-Eval, testing instances are provided in Chinese, English, and four minority languages, enabling a clear comparison between high- and low-resource languages. This parallelism also helps researchers from diverse linguistic backgrounds interpret the results.

Additionally, MiLiC-Eval demonstrates task parallelism by using the same text for multiple tasks. This cross-task parallelism enables a more faithful evaluation of LLMs, reducing the reliance on shortcuts or superficial patterns.
We report the parallelism of tasks in Table~\ref{tab:dataset_statistics}.

\paragraph{Fine-Grained Skill Evaluation}
Existing multilingual benchmarks often prioritize task quantity, with limited attention to the relationships between tasks or the skills they assess~\cite{ahuja-etal-2023-mega, zhang2024p}.

As illustrated in Figure~\ref{fig:task_skills}, MiLiC-Eval systematically formulates the tasks to assess a broad range of skills, categorized into two domains: linguistic skills, which reflect various levels of language proficiency, and problem-solving skills, which assess the ability to tackle real-world challenges. For instance, the vocabulary understanding task evaluates lexical knowledge, while the math reasoning task involves understanding the semantics of the question and performing symbolic reasoning.

For certain task types, MiLiC-Eval provides multiple subsets for more fine-grained evaluation, such as the separate machine translation subsets for translating formal texts and colloquial dialogues.

\subsection{Tasks and Dataset Curation}

\begin{figure}[t]
\centering
  \includegraphics[width=\columnwidth]{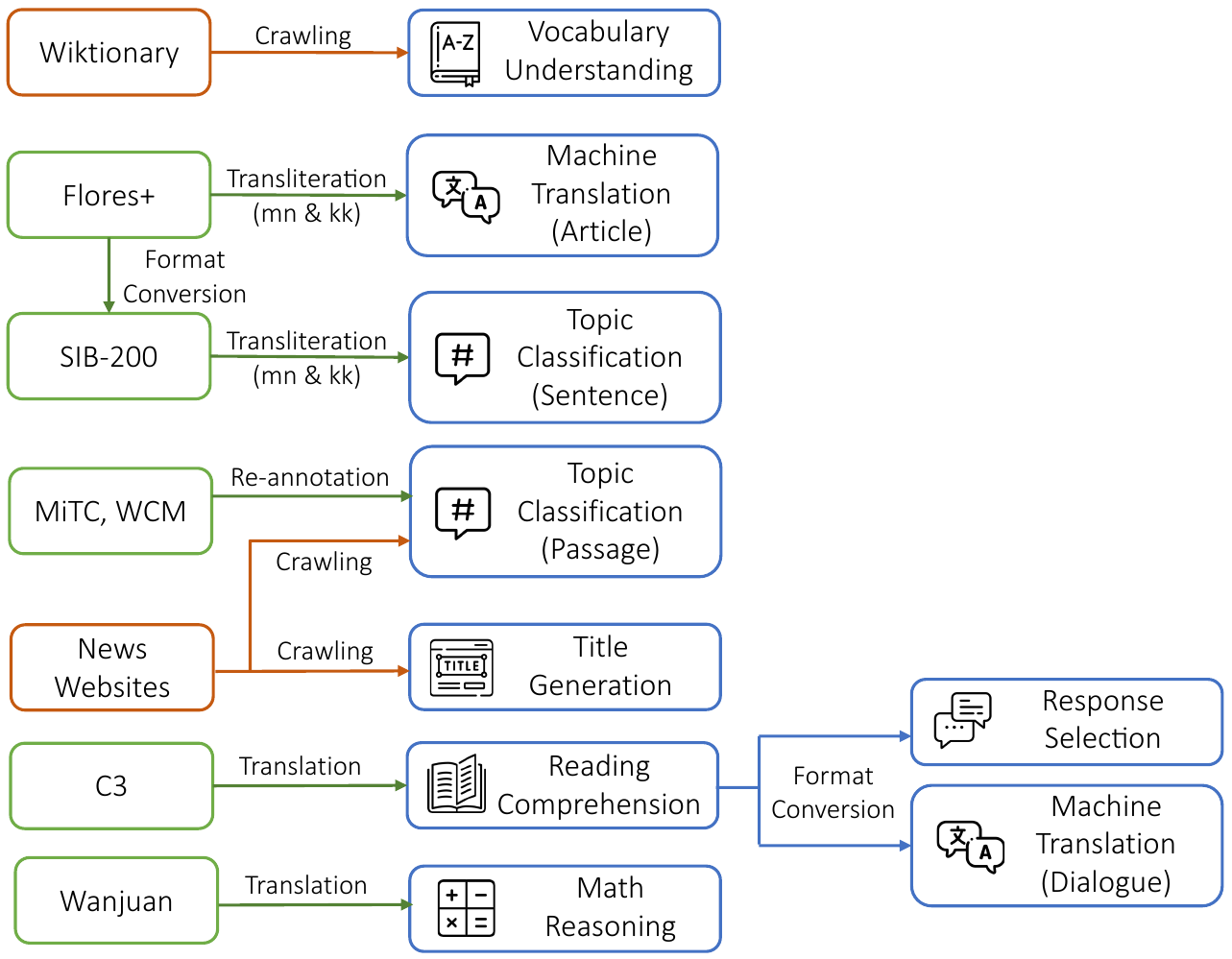}
  \caption{The methods of dataset construction for each task in MiLiC-Eval.}
  \label{fig:construction_relation}
\end{figure}

Here we introduce the tasks in MiLiC-Eval and describe the procedure of dataset curation. In Figure~\ref{fig:construction_relation}, we illustrate how each task in MiLiC-Eval is constructed. See details of dataset collection in Appendix~\ref{appendix:annotation}.

\paragraph{Vocabulary Understanding}
Accurately interpreting words is a prerequisite for language comprehension. Although straightforward, vocabulary understanding in MiLiC-Eval specifically evaluates whether LLMs truly grasp the lexicons of LRLs.
This task presents multi-choice questions where LLMs must select the correct meaning of a given word from four options. The word-meaning pairs are sourced from Wiktionary\footnote{\url{https://en.wiktionary.org/}}, with distractors sampled from words sharing the same part of speech. We construct 1,000 questions for each language.

\paragraph{Topic Classification}
Topic classification is a fundamental NLU task that tests whether LLMs can identify the topic of a given text in a minority language. Existing multilingual datasets, such as SIB-200~\cite{adelani-etal-2024-sib} and TAXI-1500~\cite{ma2023taxi1500}, have broad coverage but exclude several languages in our study. 
While datasets like WCM~\cite{yang-etal-2022-cino} and MiTC~\cite{deng2023milmo} exist for these languages, our manual audit reveals quality issues, such as errors from the automated collection,  and indistinguishable labels (e.g., \textit{culture} vs. \textit{literature}).

To address these issues, MiLiC-Eval introduces two subsets: \textbf{Sentence} and \textbf{Passage}. The \textbf{Sentence} subset, with 492 instances per language, builds upon SIB-200, which includes sentences from Wikipedia, allowing for parallel comparisons across languages. We expand this subset to include Mongolian in the traditional script and Kazakh in the Arabic script~\footnote{We transliterate the Cyrillic Kazakh instances in SIB-200 into the Arabic Kazakh script using rules. 
Since there is no exact mapping for transliteration from Cyrillic to traditional Mongolian, we recruit native speakers from Inner Mongolia to manually transliterate the Cyrillic Mongolian instances in SIB-200 into the traditional Mongolian script.}. 
The \textbf{Passage} subset, with 600 instances per language, evaluates language understanding with passages originally written in the target languages. It is reconstructed from WCM and MiTC by manually removing errors and indistinguishable labels.

\paragraph{Reading Comprehension}
Reading comprehension evaluates LLMs' multiple linguistic abilities and contextual understanding.
The dataset with the widest language coverage so far,  Belebele~\cite{bandarkar-etal-2024-belebele}, excludes three of the languages in our study.
Following Belebele, we collect a reading comprehension dataset for China's minority languages by translating 250 instances from C3~\cite{sun-etal-2020-investigating}, which are collected from Chinese exams for foreign learners.
We recruit native speakers of minority languages to translate the multi-choice instances, which consist of a dialogue, a related question, and four options.

\paragraph{Response Selection}
Response selection tests an LLM's ability to understand context and apply pragmatic reasoning to respond appropriately. 
We reuse the dialogues from our reading comprehension data to construct this task for the four minority languages.
Each response selection instance presents up to three dialogue turns, and the model must choose the most appropriate response from four options. The three distractors are the top-3 most similar sentences from other dialogues, based on SBERT-based similarity comparisons of their Chinese translations~\cite{reimers-gurevych-2020-making}. We collect 507 instances for each language.

\paragraph{Title Generation}
Title generation task evaluates LLMs' ability to generate coherent and concise texts in the target languages after understanding longer documents.
Following the approach of XLSUM~\cite{hasan-etal-2021-xl}, a multilingual summarization dataset, we collect title-article pairs from news websites originally written in minority languages. To avoid contamination from the MC$^2$ corpus~\cite{zhang-etal-2024-mc2}, which includes news texts until November 2023, we ensure the articles are published after March 2024. In total, we collect 1,000 instances per language.

\paragraph{Machine Translation}
Machine translation (MT) enables communication across languages, with Flores+~\cite{nllb-24} offering the widest language coverage. However, it lacks Kazakh in Arabic script and Mongolian in traditional Mongolian script, and mainly includes formal texts from Wikipedia, overlooking colloquial language use.
MiLiC-Eval addresses these gaps with two MT subsets: \textbf{Article} and \textbf{Dialogue}. The \textbf{Article} subset extends Flores+ to include all four languages in our study. 
We reuse the translated instances in the Topic Classification (Sentence) task as it uses the same set of texts as Flores+.
The \textbf{Dialogue} subset assesses MT performance on colloquial texts using dialogues from the reading comprehension task.
We split the dialogues into sentences, resulting in 773 instances per language.

\paragraph{Math Reasoning}
Solving math problems demonstrates LLMs' ability for complex reasoning. However, the widely-used MGSM dataset~\cite{shi2023language} does not include the languages in our study. To fill this gap, we create the first math reasoning dataset for these languages by translating 250 primary school-level math problems from Wanjuan~\cite{he2023wanjuan} into the four minority languages with native speakers' help.
As noted by \citet{shi2023language}, LLMs perform best with chain-of-thought (CoT) reasoning in high-resource languages. 
To ensure compatibility with this best practice, we manually create CoT explanations in English and Chinese for the questions in the training and development sets.

\subsection{Evaluation}

\paragraph{Metrics}
The metrics for each task are listed in Table~\ref{tab:dataset_statistics}, consistent with those used in previous works~\cite{ruder-etal-2023-xtreme,asai-etal-2024-buffet}.

\paragraph{In-Context Learning}
MiLiC-Eval is a benchmark designed specifically for the evaluation of LLMs. 
Unlike traditional supervised learning, LLMs require only a few in-context learning (ICL) examples to perform tasks effectively.
For each task, we provide a small set of instances as ICL examples, forming the training set. Three training sets are provided for each task, and we recommend running experiments multiple times with different sets to reduce variance.

\paragraph{Skill-wise Evaluation} 
As shown in Figure~\ref{fig:task_skills}, MiLiC-Eval labels each task with specific skills it aims to assess, allowing for skill-wise model evaluation.
We calculate the score of a skill by averaging the scores of tasks that assess this skill.

\begin{figure*}[t]
\centering
  \includegraphics[width=1.9\columnwidth]{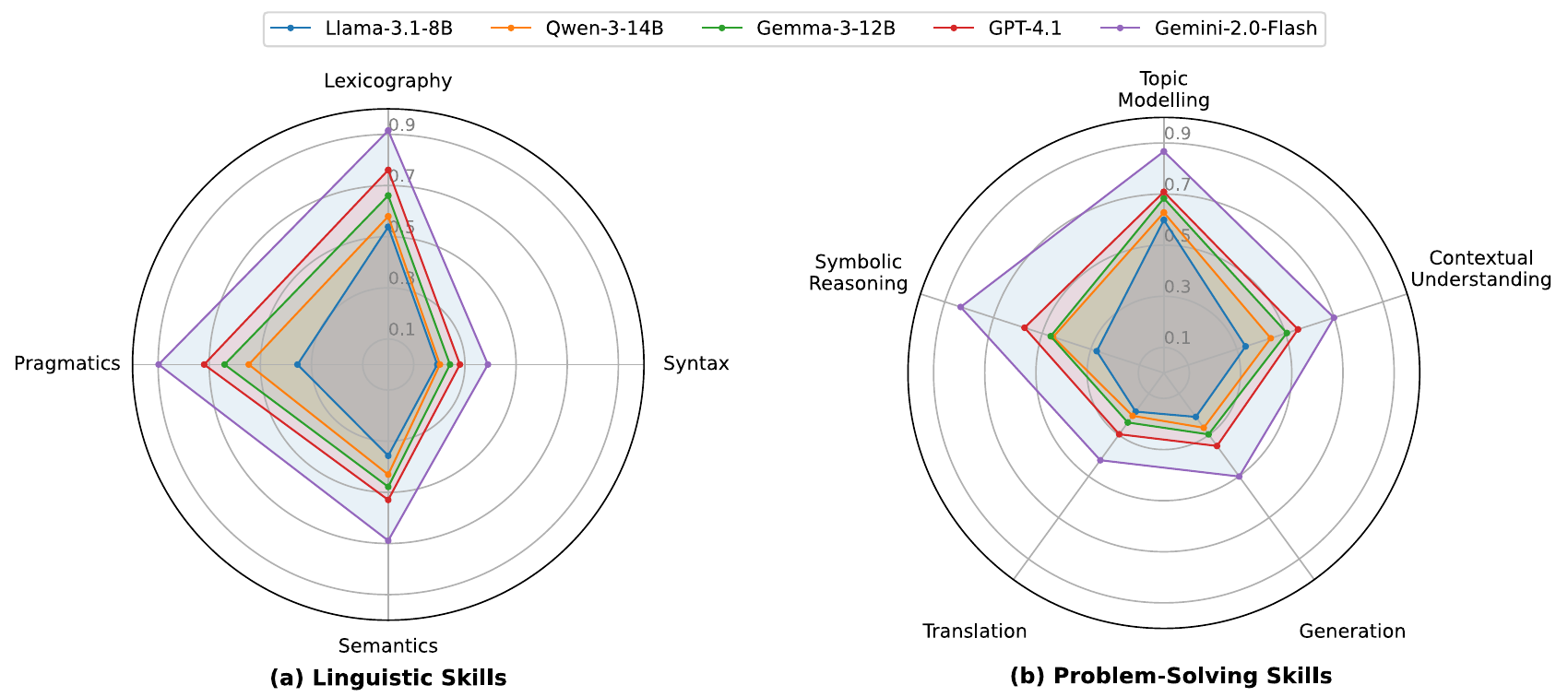}
  \caption{Skill-wise scores of the best-performing LLMs on MiLiC-Eval. We report the average scores on the four minority languages. }
  \label{fig:radar}
\end{figure*}

\section{Evaluating Multilingual LLMs}
We conduct a systematic evaluation of existing multilingual LLMs with MiLiC-Eval and analyze the current progress in China's minority languages.

\label{sec:eval}
\subsection{Experimental Setups}
\paragraph{Evaluated LLMs}
We evaluate a series of competitive proprietary and open-source LLMs.
The evaluated proprietary LLMs include \textbf{GPT-4o-mini}~\cite{hurst2024gpt}, \textbf{GPT-4.1}~\cite{gpt4.1} and \textbf{Gemini-2.0-Flash}~\cite{gemini2.0}.
Regarding open-source ones, we evaluate two sets of LLMs: (1) \textit{native multilingual LLMs}, which acquire their multilingual abilities through pretraining, including \textbf{Llama-3.1}~\cite{dubey2024llama}, \textbf{Aya-Expanse}~\cite{dang2024aya}, \textbf{Gemma-2}~\cite{team2024gemma}, \textbf{Gemma-3}~\cite{Kamath2025Gemma3T}, \textbf{Qwen-2.5}~\cite{yang2024qwen2}, \textbf{Qwen-3}~\cite{yang2025qwen3}, and \textbf{Ministral}~\cite{ministral}; (2) \textit{multilingually-adapted LLMs}, whose multilingual capabilities are enhanced via multilingual continual pretraining, including \textbf{EMMA-500}~\cite{ji2024emma500}, \textbf{BayLing-2}~\cite{zhang2024bayling}, \textbf{LLaMAX-3}~\cite{lu-etal-2024-llamax}, and \textbf{TowerInstruct}~\cite{alves2024tower}.

Among these models, only EMMA-500 and BayLing-2 explicitly claim to use training data for the minority languages of our study.
We mainly evaluate the version with around 10B parameters for open-source models. Except for EMMA-500, all the evaluated models are instruction-tuned.
Additionally, we evaluate all available versions within the Qwen-2.5 series to investigate the impact of model size. See analyses regarding the effects of model sizes in Appendix~\ref{appendix:model_sizes}.

\paragraph{Implementation Details}
Following previous works~\cite{shi2023language,asai-etal-2024-buffet}, we use English as the prompting language.
For LLMs with fewer than 10B parameters, we run the experiments three times and report the mean results. For LLMs larger than 10B and proprietary LLMs, we only run the experiments once for efficiency.
See details in Appendix~\ref{appendix:implementation}.

\begin{table}[t]
\small
\centering
\begin{tabular}{l|cccc|c}
\toprule
\textbf{Model} &  \textbf{\texttt{bo}} & \textbf{\texttt{ug}} & \textbf{\texttt{kk}} & \textbf{\texttt{mn}} & \textbf{Avg.}\\
\midrule
Ministral-8B & 21.7 & 36.0 & 32.5 & 21.7 & 28.0 \\
Aya-Expanse-8B & 22.5 & 39.6 & 39.1 & 20.5 & 30.4\\
Llama-3.1-8B & 41.3 & 52.5 & 36.4 & 20.8 &  37.8\\
Qwen-2.5-7B & 29.4 & 48.0 & 37.0 &  24.9 & 34.8\\
Qwen-3-8B & 34.5 & 56.5 & 46.7 & 28.7 & 41.6 \\
Qwen-3-14B & 41.2 & 60.0 & 50.4 & 27.2 & 44.7 \\
Gemma-2-9B & 46.9 & 53.8 & 40.6 & 25.0 & 41.6 \\
Gemma-3-12B & 53.3 & 63.7 & 57.5 & 25.1 & 49.9 \\
\midrule
TowerInstruct-7B & 17.1 & 26.1 & 26.1 & 6.5 & 19.0 \\
EMMA-500-7B & 25.3 &  42.5 & 27.4 & 17.8 &  28.2\\
BayLing-2-8B & 28.1 & 41.2 & 38.6 & 7.6 & 28.9\\
LLaMAX-3-8B & 25.2 & 43.6 & 31.0 & 18.8 & 29.7\\
\midrule
GPT-4o-mini & 36.6 & 63.6 & 48.9 & 20.7 & 42.4 \\
GPT-4.1 & \underline{57.0} & \underline{72.0} & \underline{65.9} & \underline{27.2} & \underline{55.5} \\
Gemini-2.0-Flash & \textbf{72.9} & \textbf{75.0} & \textbf{70.9} & \textbf{66.8} & \textbf{71.4} \\
\bottomrule
\end{tabular}
\caption{Language-wise scores of the evaluated LLMs on MiLiC-Eval. We report the average scores on all the tasks. The highest scores are indicated in \textbf{bold}, while the second-highest scores are marked with \underline{underline}. }
\label{tab:performance_across_lang}
\end{table}

\subsection{Results}
\label{sec:results}

We present the average performance across all tasks for each language in Table~\ref{tab:performance_across_lang}. 
Skill-wise performance for the best-performing LLMs is depicted in Figure~\ref{fig:radar}. 
For detailed scores, see Appendix~\ref{appendix:full_results}.

\paragraph{Unbalanced Performance Across Languages}
As shown in Table~\ref{tab:performance_across_lang}, despite being trained on the data in several China's minority languages, the multilingually-adapted LLMs EMMA-500 and BayLing-2 do not achieve the highest performance among the evaluated models.
Interestingly, most native multilingual LLMs exhibit a certain degree of understanding across the four minority languages, even though none explicitly claim support for these languages.
However, their performance on these languages remains significantly lower than that on high-resource languages. 
For example, the performance of Qwen-2.5-7B on Uyghur, which has the best performance among the four minority languages, is only 59\% of that on Chinese and 64\% of that on English.
See detailed comparison with Chinese and English in Table~\ref{tab:comparison_with_hrl} of the Appendix.

Moreover, the performance is severely unbalanced across the four LRLs. As shown in Table~\ref{tab:performance_across_lang}, current open-source LLMs exhibit a decent understanding of Uyghur, and a preliminary understanding of Tibetan and Kazakh, but struggle with comprehending Mongolian. 
Even the best-performing open-source model, Gemma-3, only achieves performance slightly better than random on several Mongolian tasks.
The low performance of minority languages is partially attributed to their underrepresented writing systems, which we will discuss further in Section~\ref{sec:writing_systems}.
Still, it is encouraging to see that proprietary LLMs, especially Gemini-2.0-Flash, show competitive support on the four languages. We hope that this progress could be transferred to open-source models in the future.

\paragraph{Disparities of Skills}
As illustrated in Figure~\ref{fig:radar}, most LLMs exhibit disparities across various levels of skills in LRLs. 
Regarding linguistic competence, LLMs demonstrate an ability to comprehend the semantics of input texts by recognizing the meanings of a limited set of words. 
Furthermore, they show preliminary proficiency in pragmatics and interactive communication. 
However, their grasp of syntax appears to be shallow, as they struggle to output coherent and grammatically correct sentences in the target languages.

From a problem-solving perspective, current LLMs demonstrate the ability to model the topics of texts in LRLs at a reasonable level. 
However, their capacity for fine-grained contextual understanding is still limited. 
Additionally, LLMs face great challenges in both generation and translation tasks. 
Regarding symbolic reasoning, LLMs can leverage English CoT to successfully tackle a subset of math problems despite the limited linguistic abilities in these LRLs.
% Interestingly, we observe that using English for CoT helps LLMs solve math problems in LRLs. 

\section{Discussion: What MiLiC-Eval Offers for LRL Research}
Here we discuss what MiLiC-Eval can offer for LRL research.
First, MiLiC-Eval provides a valuable resource for studying the multiplicity of writing systems, offering insights into how LLMs handle typologically unique scripts.
Second, MiLiC-Eval ensures more faithful evaluations of model performance with its cross-task parallelism and human-translated data.
Third, MiLiC-Eval, with its fine-grained task line-up, allows for a deeper understanding of the mechanisms of LLM learning new languages.

\begin{figure}[t]
\centering
  \includegraphics[width=\columnwidth]{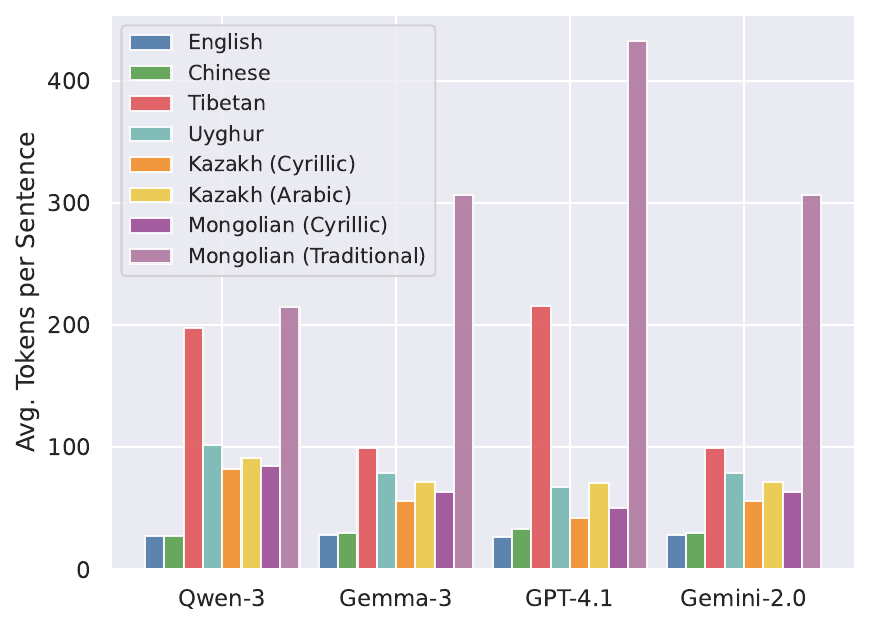}
  \caption{Average token counts of sentences in various languages from the MiLiC-Eval Machine Translation (Article) task, processed by different LLM tokenizers.}
  \label{fig:fertility}
\end{figure}

\subsection{Underrepresented Writing Systems}
\label{sec:writing_systems}
LLMs are typically trained on corpora dominated by English, with non-Latin scripts receiving limited attention. 
This neglect can lead to poor performance, higher computational costs, and the marginalization of less-represented cultures~\cite{ahuja-etal-2023-mega, ahia-etal-2023-languages, zhang-etal-2024-mc2}.
Through MiLiC-Eval, we observe that current LLMs struggle with underrepresented scripts, showing inefficiencies in tokenization and incompatibilities of writing systems. 
These challenges emphasize the need for improved tokenization strategies and robust handling of multiple scripts.

\paragraph{Inefficiency in Tokenization} 
We examine tokenization efficiency using parallel sentences from the Machine Translation (Article) task in MiLiC-Eval.
We calculate the average number of tokens required to encode sentences in different languages or scripts.
As shown in Figure~\ref{fig:fertility}, the average token count for English (27 tokens) and Chinese (31 tokens) is much lower compared to the four minority languages, particularly Tibetan and traditional Mongolian, which require 100-430 tokens per sentence. 
For languages that employ multiple writing systems, tokenization efficiency for less common scripts is often considerably lower than for more widely used scripts. For example, GPT-4.1 requires 432 tokens for a sentence in traditional Mongolian, eight times the number of tokens needed for the Cyrillic script.

\paragraph{Incompatibility of Multiple Writing Systems}
We observe that current LLMs often exhibit code-switching errors when generating content in minority languages, i.e., switching to other languages or scripts during generation, which is also known as language confusion~\cite{marchisio-etal-2024-understanding}. 
This issue is especially prevalent in the Kazakh and Mongolian languages used in China, both of which employ less commonly used writing systems. For instance, in the title generation task, the GPT-4o-mini model switches to Cyrillic, the more widely used script for both languages, in 36\% of Mongolian cases and 95\% of Kazakh cases. 
This problem significantly hurts model performance on generation tasks and may lead to confusion for users when deployed in real-world applications.

\subsection{Faithful Evaluation of LRL Abilities}

Prior studies on LRLs often rely on a single task format for evaluation, such as simple NLU tasks~\cite{yong-etal-2023-bloom, luukkonen-etal-2023-fingpt, lin2024mala} or translation~\cite{nllb-24}. 
However, such an over-reliance on a single task type may lead to biased assessments of model capabilities. 
Additionally, many studies use machine-translated data for evaluation~\cite{hu2020xtreme, chen-etal-2024-breaking,huang2025benchmax}, but the inherent noise in such data can obscure the true performance of models. 
In contrast, MiLiC-Eval uses diverse task formats derived from the same set of documents and recruits native speakers to translate the data.
We discuss how these designs can provide a more robust and faithful assessment of model performance in LRLs.

\begin{figure}[t]
\centering
  \includegraphics[width=\columnwidth]{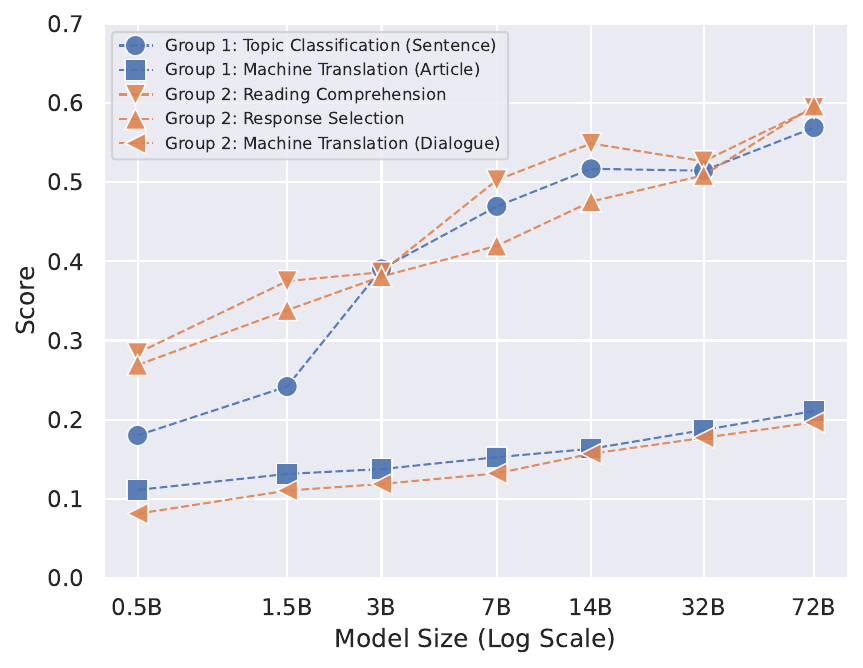}
  \caption{The relationship between model sizes and the scores on tasks of different formats. }
  \label{fig:task_format}
\end{figure}

\paragraph{Task Parallelism}
We argue that reliance on a single evaluation task can result in a skewed assessment of a model’s true capabilities in LRLs. 
To demonstrate, we select two task groups from MiLiC-Eval, where the tasks within each group are derived from the same set of texts. 
We evaluate these tasks across different sizes of Qwen-2.5. The results are presented in Figure~\ref{fig:task_format}.

As the models' overall capabilities increase with the scaling of sizes, the speed of improvement varies significantly across different task formats. 
NLU tasks, such as topic classification, improve rapidly, while translation tasks show a much slower rate of improvement.
This indicates that evaluation on simple NLU tasks only might lead to over-optimistic conclusions, while solely using translation tasks for evaluation might underestimate models' abilities. 
In contrast, by diversifying task formats through task parallelism, MiLiC-Eval can provide a more faithful and comprehensive assessment of LLMs’ understanding of LRLs.

\paragraph{Human Translation}
To evaluate the effect of using MT systems for data construction, we use NLLB-200-3.3B~\cite{nllb-24}, a widely-used MT model, to translate several tasks in MiLiC-Eval from Chinese into minority languages\footnote{Note that despite its wide language coverage, NLLB-200 does not support the traditional Mongolian script.}.
Table~\ref{tab:mt_data} shows the model performance on machine-translated task data.

We observe a consistent performance drop with machine-translated data compared to human-translated data, particularly on the math reasoning task, which demands a more accurate translation. The drop is more pronounced in languages with poor MT support, such as Tibetan. 
The chrF++ scores of translation results are 23.0 for Tibetan, 30.4 for Uyghur, and 26.3 for Kazakh.
Tibetan has the lowest chrF++ score among the three languages, which also exhibits the most significant performance decline resulting from translated evaluation data.

We further check the translation results of NLLB0-200 and find that the most commonly observed error is repetition, where the model keeps outputting the same words~\cite{wang-etal-2024-mitigating-language}.
Additionally, the model often omits or modifies numerical values when translating math problems.
These findings underscore the importance of native speakers in collecting evaluation data for LRLs.

\begin{table}[t]
\small
\centering
\begin{tabular}{l|ccc}
\toprule
\textbf{Language} & \textbf{Reading} & \textbf{Response} & \textbf{Math} \\
\midrule
Tibetan & 40.0 \footnotesize{(-21\%)} & 36.9 (-15\%) & 11.9 (-52\%) \\
Uyghur & 41.8 (-19\%) & 42.2 (-17\%) & 31.3 (-29\%)\\
Kazakh & 40.3 (-19\%) & 32.3 (-16\%) & 19.3 (-22\%) \\
\bottomrule
\end{tabular}
\caption{Average scores (\%) of three LLMs (Qwen-2.5, Llama-3.1, and Gemma-2) on the task data constructed by NLLB-200-3.3B. The number in parentheses represents the percentage of decrease in performance relative to that evaluated on human-translated data.}
\label{tab:mt_data}
\end{table}

\begin{figure}[t]
\centering
  \includegraphics[width=0.85\columnwidth]{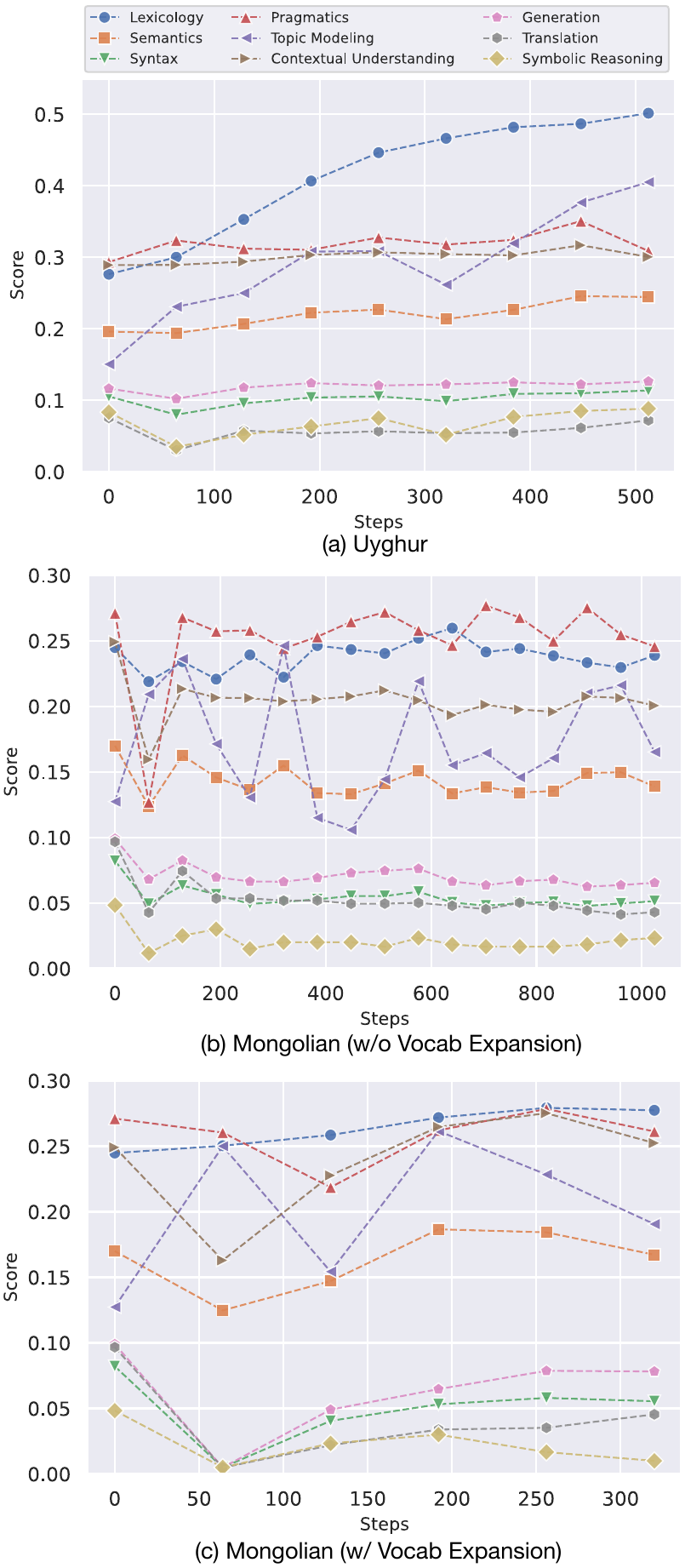}
  \caption{Tracking the process of continual pretraining with MiLiC-Eval on Uyghur and Mongolian.}
  \label{fig:ct}
\end{figure}

\subsection{Skill-wise Tracking of Language Adaptation}
MiLiC-Eval consists of a wide range of tasks assessing different skills.
Through these tasks, we can track how a model's abilities evolve as it learns a new language, providing deeper insights into various language adaptation techniques.

We take continual pretraining as an example, a widely used approach for language adaptation, and examine how a model's capabilities develop in target languages. Specifically, we continually pretrain Qwen-2.5-0.5B on the MC$^2$ corpus~\cite{zhang-etal-2024-mc2} for Uyghur and Mongolian. These two languages represent the highest and lowest performance points for Qwen-2.5-0.5B among the four minority languages in our study. 
See implementation details in Appendix~\ref{appendix:implementation}.

Figure~\ref{fig:ct} presents the continual pretraining course of the two languages. 
For Uyghur, to which Qwen-2.5 has had some exposure during pretraining, we observe a consistent improvement across all evaluated abilities. In particular, lexicography and topic modeling show the most significant gains, whereas abilities such as syntax, generation, translation, and symbolic reasoning exhibit modest or little improvement.

In contrast, Qwen-2.5 has limited exposure to Mongolian, and its tokenizer exhibits poor support, tokenizing most Mongolian characters as bytes. 
Consequently, continual pretraining yields minimal improvements on Mongolian. However, after expanding the tokenizer by 3K tokens obtained from the pretraining corpus~\cite{hewitt2021initializing}, we observe noticeable improvement in several abilities for Mongolian, especially in lexicography and topic modeling, similar to the patterns observed on Uyghur.
This finding challenges previous studies claiming that vocabulary expansion has little effect on downstream performance~\cite{csaki2024sambalingo}. 
We also note an initial decline in performance for certain Mongolian abilities at the beginning of training after vocabulary expansion. 
This drop may be due to the model's adaptation to the newly added vocabulary, after which performance improves as the model adjusts to the expanded lexicon.

Our findings underscore the necessity of rethinking the common practice of continual pretraining, which may have limitations in enhancing certain abilities for LRLs, such as generation and translation. 
Furthermore, the trade-offs in different design choices, such as vocabulary extension, can significantly impact the training process. 
We hope that MiLiC-Eval, through its fine-grained evaluation methodology, can serve as a diagnostic tool for language adaptation techniques, facilitating a deeper understanding of how LLMs learn new languages.

\section{Related Works}
\paragraph{NLP for Minority Languages in China}
Previous research has focused on improving the accessibility of China's minority languages by collecting resources including pretraining corpora~\cite{zhang-etal-2024-mc2,zhuang-sun-2025-cute} and task-specific datasets such as topic classification~\cite{qun2017end,yang-etal-2022-cino,deng2023milmo}, question answering~\cite{sun2021teaching,bandarkar-etal-2024-belebele}, and machine translation~\cite{nllb-24,zhang-etal-2024-teaching}.
However, there is no standardized benchmark to track the progress of LLMs in these languages. MiLiC-Eval is the first large-scale multi-task benchmark to address this gap, aiming to facilitate research for these languages.

\paragraph{Multilingual Evaluation of LLMs}
Recent works on multilingual evaluation primarily focus on generative abilities~\cite{ahuja-etal-2023-mega,singh-etal-2024-indicgenbench}, user-centric tasks~\cite{ruder-etal-2023-xtreme}, cross-lingual transfer~\cite{asai-etal-2024-buffet}, and cultural awareness~\cite{wang-etal-2024-seaeval,romanou2024include}. 
However, these benchmarks are usually simple combinations of existing datasets and have limited coverage of LRLs, particularly those examined in our work. 
In contrast, MiLiC-Eval, with systematic task line-ups, offers a comprehensive and faithful evaluation for LRLs and supports multiple research directions.

\section{Conclusion}
We present MiLiC-Eval, a multilingual benchmark comprising 24K instances across 9 tasks and 4 minority languages in China. MiLiC-Eval is distinguished by three key features: (1) a focus on underrepresented writing systems, (2) cross-language and cross-task parallelism, and (3) fine-grained skill-wise evaluation. 
We hope that MiLiC-Eval will not only advance LLM support for LRLs in China but also inspire research on underrepresented languages in other regions, such as Africa, India, and Southeast Asia.

\section*{Limitations}
\paragraph{Translation-based Collection} 
Several tasks in MiLiC-Eval are translated by native speakers from datasets in high-resource languages. This approach may introduce biases or lead to translationese. To mitigate these issues, MiLiC-Eval also includes tasks sourced directly from native texts in each target language, such as text classification (passage) and title generation.

\paragraph{Culture-related Tasks}
A key challenge in developing multilingual LLMs is enhancing their cultural awareness, particularly for underrepresented cultures. 
MiLiC-Eval primarily evaluates the linguistic and problem-solving abilities of LLMs in the four minority languages of China.
It does not include tasks that specifically assess cultural knowledge tied to the cultures behind these languages.
We believe the collection of culture reasoning tasks as future work.

\section*{Acknowledgements}
This work is supported in part by NSFC~(62161160339) and Beijing Science and Technology Program~(Z231100007423011). We thank the anonymous reviewers for their valuable suggestions. 
We also thank all the annotators who contributed to the data collection.
For any correspondence, please contact Yansong Feng.

\bibliography{anthology,custom}

\clearpage
\appendix

\section{Information of Studied Languages}
In Table~\ref{tab:language_info}, we report basic information about the minority languages of our study.

\section{Details of Data Collection}

\label{appendix:annotation}
\begin{table*}[t]
\small
\centering
\begin{tabular}{c|cc|c|c}
\toprule
\textbf{Name} & \textbf{ISO 639-1} & \textbf{ISO 639-3} & \textbf{Language Family} & \textbf{Writing System}  \\
\midrule
Tibetan & \texttt{bo} & \texttt{bod} & 	
Sino-Tibetan & Tibetan script \\
Uyghur & \texttt{ug} & \texttt{uig} & Turkic &  Uyghur Arabic script\\
Kazakh & \texttt{kk} & \texttt{kaz} & Turkic & Kazakh Arabic script \\
Mongolian & \texttt{mn} & \texttt{mvf} & Mongolic & Traditional Mongolian script \\
\bottomrule
\end{tabular}
\caption{ISO codes, language families, and writing systems of the languages in MiLiC-Eval.}
\label{tab:language_info}
\end{table*}

\subsection{Collection of Topic Classification (Passage)}
\label{appendix:topic_classification}
For each language, we select four labels from existing datasets according to the following criteria:
\begin{itemize}
\item Discard categories with too few instances.
\item Discard instances that include English translations of entities in the text, as this could leak the answer.
\item Discard instances that are not written in the target language.
\end{itemize}

We then select four labels for each language, which are easy to distinguish. 
We show the selected labels of each language in Table~\ref{tab:classification_label}.
Because there are not enough qualified labels for Uyghur, we additionally collect articles from a Uyghur news website \texttt{nur.cn}. We obtain 150 instances from the \textit{Health} and \textit{Sport} columns, respectively.

\subsection{Human Annotation}
\paragraph{Translation into Minority Languages}
The data collection of several tasks in MiLiC-Eval involves translating testing instances from Chinese into minority languages.
We recruit volunteers from universities who are native speakers of the four minority languages and proficient in Chinese.  
They are informed of how the collected data will be used.
They are paid approximately \$1.5 for translating a short dialogue in C3 and \$0.5 for translating a math problem in Wanjuan, which is adequate given the participants’ demographic.
Each translated instance is verified by another annotator to ensure quality.

\paragraph{Translation into English}
To obtain the English version of the tasks, we use GPT-4o-mini to translate the dialogues in C3 and the math problems from Chinese and English. Then several authors, who are proficient in both Chinese and English, check the translation results.

\paragraph{Transliteration of Mongolian}
The Mongolian instances in Flores+ are written in Cyrillic, but for our study, we need to transliterate them into traditional Mongolian, as used by the Mongolian communities in China. Since there are no strict one-to-one rules for transliterating Mongolian, we first use online tools\footnote{\url{http://trans.mglip.com/}} for the transliteration process, followed by post-editing by native speakers. They are compensated \$0.5 per sentence for post-editing.

\begin{table}[t]
\small
\centering
\begin{tabular}{lcc}
\toprule
\textbf{Language} & \textbf{Label} & \textbf{Source Dataset} \\
\midrule
\multirow{4}*{Tibetan} &  Education & MiTC \\
& Travel & MiTC \\
& Law & MiTC \\
& Economy & MiTC \\
\midrule
\multirow{4}*{Uyghur} & Livelihood & MiTC \\
& Travel & MiTC \\
& Health & Newly Collected \\
& Sport & Newly Collected \\
\midrule
\multirow{4}*{Kazakh} & Politics & MiTC \\
& Economy & MiTC \\
& Culture & MiTC \\
& Geography & WCM \\
\midrule
\multirow{4}*{Mongolian} & Health & MiTC \\
& Politics & MiTC \\
& Education & MiTC \\
& Technology & WCM \\
\bottomrule
\end{tabular}
\caption{The labels and sources of the instances in topic classification (Passage).}
\label{tab:classification_label}
\end{table}

\begin{table}[t]
\small
\centering
\begin{tabular}{lc}
\toprule
\textbf{Series} & \textbf{Used Checkpoints} \\
\midrule
Ministral & \texttt{Ministral-8B-Instruct-2410} \\
Aya-Expanse & \texttt{aya-expanse-8b} \\
Qwen-2.5 & \texttt{Qwen2.5-7B-Instruct} \\
Qwen-3 & \texttt{Qwen3-8B}, \texttt{Qwen3-14B} \\ 
Llama-3.1 & \texttt{Llama-3.1-8B-Instruct} \\
Gemma-2 & \texttt{gemma-2-9b-it} \\
Gemma-3 & \texttt{gemma-3-12b-it} \\
\midrule
TowerInstruct & \texttt{TowerInstruct-Mistral-7B-v0.2} \\
EMMA-500 & \texttt{emma-500-llama2-7b} \\
BayLing-2 & \texttt{bayling-2-llama-3-8b}\\
LLaMAX-3 & \texttt{LLaMAX3-8B-Alpaca}\\
\midrule
GPT-4o & \texttt{gpt-4o-mini-2024-07-18}\\
GPT-4.1 & \texttt{gpt-4.1-2025-04-14}\\
Gemini-2.0 & \texttt{gemini-2.0-flash-001}\\
\bottomrule
\end{tabular}
\caption{The model variants used in evaluation.}
\label{tab:model_variants}
\end{table}

\section{Implementation Details}
\label{appendix:implementation}
\paragraph{Models Used in Evaluation}
Several LLMs have multiple variants available. Here we report the variants used in our evaluation in Table~\ref{tab:model_variants}.

\paragraph{In-Context Learning} 
For the tasks in MiLiC-Eval, we use 5-shot examples, except for Title Generation, for which we use 3-shot examples due to the long lengths of news articles. 

\paragraph{Continual Pretraining} 
To mitigate the catastrophic forgetting of LLMs' English capabilities, we incorporate English data from C4~\cite{raffel2020exploring}, amounting to 20\% of the size of the target language corpus.
We use Deepseed\footnote{\url{https://github.com/microsoft/DeepSpeed}} for training. 
The model is trained for one epoch using a batch size of 0.5M tokens, a learning rate of 1e-4, and a warmup ratio of 0.01 on eight A100 GPUs.
A training step takes approximately 18 seconds.

\section{Additional Experiment Results}
\subsection{Full Evaluation Results on MiLiC-Eval}
\label{appendix:full_results}
In Table~\ref{tab:main_experiment_bo}, Table~\ref{tab:main_experiment_ug}, Table~\ref{tab:main_experiment_kk}, and Table~\ref{tab:main_experiment_mn}, we report the scores of each model on each task in MiLiC-Eval. 

Additionally, we report the performance of Qwen-2.5-7B, Llama-3.1-8B, and Gemma-2-9B on the Chinese and English versions of the tasks in Table~\ref{tab:performance_across_lang}.

\begin{figure}[t]
\centering
  \includegraphics[width=0.9\columnwidth]{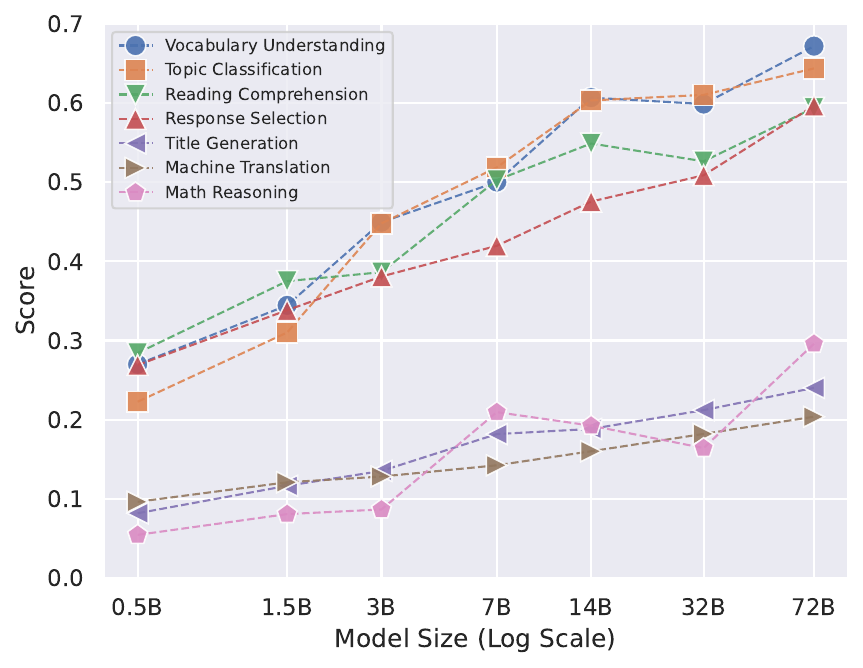}
  \caption{Performance of Qwen-2.5 across different model sizes on the MiLiC-Eval dataset. The reported scores are averaged across the four minority languages.}
  \label{fig:model_size}
\end{figure}

\subsection{Effect of Model Sizes}
\label{appendix:model_sizes}
We evaluate LLM performance across different model sizes in the Qwen-2.5 series, ranging from 0.5B to 72B parameters. As shown in Figure~\ref{fig:model_size}, performance improves in a log-linear fashion with increasing model size, roughly doubling or tripling from 0.5B to 72B. However, the improvement varies by task: basic NLU tasks like vocabulary understanding and topic classification show the largest gains, while generation tasks and math reasoning remain challenging. This indicates that simply scaling up sizes might not be the best practice for LRLs.

\begin{table*}[t]
\small
\centering
\setlength\tabcolsep{5pt}
\begin{tabular}{l|ccccccccccc|c}
\toprule
\textbf{Model} &  \textbf{{Vocab.}} & \textbf{{TC}} & \textbf{{TC}} & \textbf{{Read.}} & \textbf{Resp.} & \textbf{Title} & \textbf{MT-A} & \textbf{MT-A} & \textbf{MT-D} & \textbf{MT-D} & \textbf{Math} & \textbf{Avg.}\\
 & \textbf{Test} & \textbf{Sent.} & \textbf{Pass.} & \textbf{Comp.} & \textbf{Sel.} & \textbf{Gen.} & xx2en & en2xx & xx2en & en2xx &  & \\
\midrule
 Random & 25.0 & 16.2 & 25.0 & 27.0 & 25.0 & - & \ \ - & \ \ - & \ \ - & \ \ - & \ \ - & - \\
 \midrule
Ministral-8B & 33.1 & 22.4 & 29.9 & 33.7 & 27.3 & 23.8 & 15.7 & \ \ 5.9 & 10.2 & \ \ 8.5 & \ \ 5.3 & 21.7 \\
Aya-Expanse-8B & 32.2 & 20.6 & 28.3 & 40.3 & 28.7 & 21.3 & 16.3 & \ \ 8.0 & 13.1 & \ \ 9.5 & \ \ 7.5 & 22.5 \\
Llama-3.1-8B & 60.6 & 63.5 & 67.3 & 44.8 & 37.4 & 29.3 & 27.5 & \ \ 9.3 & 28.3 & 12.8 & 29.7 & 41.3 \\
Qwen-2.5-7B & 37.7 & 30.6 & 39.4 & 48.3 & 39.6 & 27.6 & 18.8 & 10.8 & 14.2 & 13.2 & 13.3 & 29.4 \\
Qwen-3-8B & 43.7 & 36.9 & 67.3 & 52.2 & 39.8 & 20.2 & 20.2 & \ \ 8.6 & 17.8 & 10.4 & 21.5 & 34.5 \\
Qwen-3-14B & 53.4 & 52.1 & 68.5 & 54.0 & 49.1 & 21.2 & 23.4 & 11.9 & 22.8 & 15.8 & 35.5 & 41.2 \\
Gemma-2-9B & 60.0 & 61.8 & 78.0 & 58.2 & 53.0 & 34.4 & 25.2 & 15.6 & 27.8 & 19.6 & 32.7 & 46.9 \\
Gemma-3-12B & 65.9 & \underline{70.8} & \underline{84.7} & 63.0 & 58.0 & \underline{36.6} & 23.7 & 20.6 & 31.4 & 23.1 & 51.0 & 53.3 \\
\midrule
TowerInstruct-7B & 26.8 & 24.1 & 28.6 & 26.0 & 25.8 & \ \ 1.2 & 13.3 & \ \ 4.9 & 11.0 & \ \ 6.2 & \ \ 4.0 & 17.1 \\
EMMA-500-7B & 47.3 & 39.4 & 48.8 & 42.5 & 27.0 & \ \ 0.2 & 21.3 & 10.4 & 25.0 & 14.4 & 12.0 & 28.1 \\
Bayling-2-8B & 43.7 & 41.4 & 39.3 & 28.5 & 26.8 & 17.9 & 15.4 & \ \ 7.8 & 14.1 & \ \ 8.1 & \ \ 7.5 & 25.3 \\
LLaMAX-3-8B & 44.0 & 34.5 & 26.0 & 45.5 & 30.5 & \ \ 7.5 & 20.7 & \ \ 5.4 & 20.0 & \ \ 5.2 & 13.5 & 25.2 \\
\midrule
GPT-4o-mini & 66.0 & 40.7 & 60.5 & 40.0 & 38.8 & 24.3 & 23.2 & 16.3 & 21.5 & 18.0 & 19.5 & 36.6 \\
GPT-4.1 & \underline{85.3} & 66.7 & 84.4 & \underline{64.0} & \underline{65.8} & 32.1 & \underline{35.4} & \underline{24.6} & \underline{38.3} & \underline{25.1} & \underline{53.0} & \underline{57.0} \\
Gemini-2.0-Flash & \textbf{91.7} & \textbf{84.5} & \textbf{92.3} & \textbf{87.0} & \textbf{89.4} & \textbf{42.1} & \textbf{48.8} & \textbf{33.4} & \textbf{53.8} & \textbf{34.2} & \textbf{84.0} & \textbf{72.9} \\
\bottomrule
\end{tabular}
\caption{Scores (\%) of different LLMs on the \textbf{Tibetan} tasks of MiLiC-Eval. 
\textbf{Vocab. Test} refers to Vocabulary Understanding. 
\textbf{TC Sent.} refers to topic classification (Sentence).
\textbf{TC Pass.} refers to topic classification (Passage).
\textbf{Read. Comp.} refers to Reading Comprehension.
\textbf{Resp. Sel.} refers to Response Selection.
\textbf{Title Gen.} refers to Title Generation.
\textbf{MT-A} refers to Machine Translation (Article).
xx2en denotes translation from the minority languages to English.
en2xx denotes translation from English to the minority languages.
\textbf{MT-D} refers to Machine Translation (Dialogue).
\textbf{Math} refers to Math Reasoning.
When calculating the average, the score for the two translation tasks is the mean of xx2en and en2xx. The highest scores are indicated in \textbf{bold}, while the second-highest scores are marked with \underline{underline}.}
\label{tab:main_experiment_bo}
\end{table*}

\begin{table*}[t]
\small
\centering
\setlength\tabcolsep{5pt}
\begin{tabular}{l|ccccccccccc|c}
\toprule
\textbf{Model} &  \textbf{{Vocab.}} & \textbf{{TC}} & \textbf{{TC}} & \textbf{{Read.}} & \textbf{Resp.} & \textbf{Title} & \textbf{MT-A} & \textbf{MT-A} & \textbf{MT-D} & \textbf{MT-D} & \textbf{Math} & \textbf{Avg.}\\
 & \textbf{Test} & \textbf{Sent.} & \textbf{Pass.} & \textbf{Comp.} & \textbf{Sel.} & \textbf{Gen.} & xx2en & en2xx & xx2en & en2xx &  & \\
\midrule
 Random & 25.0 & 16.2 & 25.0 & 27.0 & 25.0 & - & \ \ - & \ \ - & \ \ - & \ \ - & \ \ - & - \\
 \midrule
Ministral-8B & 55.9 & 61.7 & 67.1 & 35.5 & 31.0 & 14.8 & 23.2 & \ \ 8.1 & 21.1 & \ \ 8.6 & 27.7 & 36.0 \\
Aya-Expanse-8B & 57.5 & 56.0 & 77.3 & 47.5 & 37.0 & 21.5 & 26.3 & \ \ 9.8 & 21.0 & \ \ 9.2 & 26.2 & 39.6 \\
Llama-3.1-8B & 73.2 & 76.9 & 91.0 & 59.2 & 46.0 & 22.4 & 41.3 & 17.6 & 36.2 & 15.9 & 48.0 & 52.5 \\
Qwen-2.5-7B & 70.7 & 70.6 & 85.1 & 57.0 & 47.8 & 22.1 & 30.8 & 12.2 & 26.4 & 11.7 & 38.0 & 48.0 \\
Qwen-3-8B & 77.7 & 77.1 & 90.7 & 65.8 & 62.6 & 23.0 & 38.0 & 14.9 & 32.8 & 14.7 & 61.7 & 56.5 \\
Qwen-3-14B & 80.6 & 78.9 & 91.7 & 70.5 & 69.3 & 22.7 & 42.9 & 18.2 & 38.0 & 17.6 & 68.0 & 60.0 \\
Gemma-2-9B & 76.1 & 75.9 & 93.7 & 58.5 & 59.5 & 28.2 & 37.0 & 12.1 & 31.7 & 11.1 & 46.8 & 53.8 \\
Gemma-3-12B & 86.4 & 75.0 & 93.8 & 77.0 & 82.1 & 27.1 & 44.9 & 23.8 & 41.1 & 20.0 & 66.5 & 63.6 \\
\midrule
TowerInstruct-7B & 41.3 & 37.3 & 47.8 & 29.5 & 27.5 & 16.8 & 13.7 & \ \ 8.0 & 14.7 & \ \ 8.3 & 12.0 & 26.1 \\
EMMA-500-7B & 66.7 & 70.6 & 37.3 & 46.0 & 38.8 & \underline{30.3} & 28.5 & 25.6 & 37.1 & 26.5 & 22.0 & 41.2 \\
Bayling-2-8B & 65.7 & 75.0 & 78.2 & 48.5 & 30.2 & 12.4 & 31.5 & 15.8 & 30.6 & 14.9 & 26.5 & 42.5 \\
LLaMAX-3-8B & 66.7 & 71.8 & 39.8 & 53.0 & 69.4 & 16.0 & 37.2 & 14.3 & 32.3 & 12.8 & 27.5 & 43.6 \\
\midrule
GPT-4o-mini & 92.4 & 67.6 & 92.5 & 75.5 & 80.8 & 18.8 & 48.6 & 28.3 & 44.5 & 23.8 & 72.0 & 63.6 \\
GPT-4.1 & \underline{95.2} & \textbf{87.3} & \textbf{96.4} & \underline{83.5} & \underline{89.4} & 26.6 & \underline{54.1} & \underline{37.5} & \underline{51.5} & \underline{31.3} & \underline{82.5} & \underline{72.0} \\
Gemini-2.0-Flash & \textbf{97.2} & \underline{85.0} & \underline{94.6} & \textbf{88.5} & \textbf{92.1} & \textbf{32.0} & \textbf{56.2} & \textbf{48.7} & \textbf{55.2} & \textbf{39.2} & \textbf{86.0} & \textbf{75.0} \\
\bottomrule
\end{tabular}
\caption{Scores (\%) of different LLMs on the \textbf{Uyghur} tasks of MiLiC-Eval. 
\textbf{Vocab. Test} refers to Vocabulary Understanding. 
\textbf{TC Sent.} refers to topic classification (Sentence).
\textbf{TC Pass.} refers to topic classification (Passage).
\textbf{Read. Comp.} refers to Reading Comprehension.
\textbf{Resp. Sel.} refers to Response Selection.
\textbf{Title Gen.} refers to Title Generation.
\textbf{MT-A} refers to Machine Translation (Article).
xx2en denotes translation from the minority languages to English.
en2xx denotes translation from English to the minority languages.
\textbf{MT-D} refers to Machine Translation (Dialogue).
\textbf{Math} refers to Math Reasoning.
When calculating the average, the score for the two translation tasks is the mean of xx2en and en2xx. The highest scores are indicated in \textbf{bold}, while the second-highest scores are marked with \underline{underline}.}
\label{tab:main_experiment_ug}
\end{table*}

\begin{table*}[t]
\small
\centering
\setlength\tabcolsep{5pt}
\begin{tabular}{l|ccccccccccc|c}
\toprule
\textbf{Model} &  \textbf{{Vocab.}} & \textbf{{TC}} & \textbf{{TC}} & \textbf{{Read.}} & \textbf{Resp.} & \textbf{Title} & \textbf{MT-A} & \textbf{MT-A} & \textbf{MT-D} & \textbf{MT-D} & \textbf{Math} & \textbf{Avg.}\\
 & \textbf{Test} & \textbf{Sent.} & \textbf{Pass.} & \textbf{Comp.} & \textbf{Sel.} & \textbf{Gen.} & xx2en & en2xx & xx2en & en2xx &  & \\
\midrule
 Random & 25.0 & 16.2 & 25.0 & 27.0 & 25.0 & - & \ \ - & \ \ - & \ \ - & \ \ - & \ \ - & - \\
 \midrule
Ministral-8B & 46.6 & 40.3 & 57.8 & 50.2 & 33.9 & 21.0 & 20.4 & \ \ 5.3 & 14.5 & \ \ 6.5 & 19.3 & 32.5 \\
Aya-Expanse-8B & 60.7 & 64.2 & 66.5 & 50.8 & 37.8 & 17.4 & 30.4 & \ \ 1.3 & 23.1 & \ \ 1.3 & 26.5 & 39.1 \\
Llama-3.1-8B & 54.9 & 55.5 & 61.8 & 45.7 & 32.9 & 19.9 & 24.3 & 11.0 & 18.9 & 11.0 & 24.0 & 36.4 \\
Qwen-2.5-7B & 60.7 & 56.5 & 64.8 & 51.7 & 40.1 & 13.8 & 25.6 & \ \ 1.0 & 21.0 & \ \ 1.1 & 21.3 & 37.0 \\
Qwen-3-8B & 65.7 & 63.7 & 74.7 & 57.0 & 57.2 & 21.9 & 30.5 & \ \ 8.2 & 25.2 & \ \ 8.9 & 43.5 & 46.7 \\
Qwen-3-14B & 68.1 & 72.0 & 73.0 & 63.0 & 57.0 & 20.4 & 34.3 & 12.2 & 28.6 & 12.8 & 56.5 & 50.4 \\
Gemma-2-9B & 60.6 & 57.9 & 66.2 & 51.7 & 43.0 & 26.4 & 23.4 & \ \ 7.8 & 19.9 & \ \ 8.7 & 29.3 & 40.6 \\
Gemma-3-12B & 80.8 & 76.6 & \textbf{80.0} & 71.5 & 74.2 & 24.2 & 47.8 & 12.1 & 36.2 & 13.8 & 55.5 & 57.5 \\
\midrule
TowerInstruct-7B & 45.2 & 38.0 & 49.0 & 22.0 & 31.4 & 19.7 & 10.2 & \ \ 8.3 & 11.9 & \ \ 8.5 & 10.5 & 26.1 \\
EMMA-500-7B & 61.9 & 40.1 & 56.8 & 43.5 & 38.1 & \textbf{35.7} & 26.1 & \underline{27.7} & 22.9 & \textbf{27.1} & 19.5 & 38.6 \\
Bayling-2-8B & 45.0 & 54.9 & 51.2 & 25.5 & 30.5 & \ \ 4.1 & 19.5 & \ \ 5.3 & 12.0 & \ \ 4.2 & 15.0 & 27.4 \\
LLaMAX-3-8B & 47.9 & 54.4 & 40.3 & 47.0 & 34.2 & 16.6 & 24.3 & \ \ 5.8 & 18.8 & \ \ 7.0 & 11.0 & 31.0 \\
\midrule
GPT-4o-mini & 79.8 & 63.4 & 73.4 & 63.5 & 61.4 & \ \ 6.0 & 40.0 & \ \ 1.3 & 34.7 & \ \ 1.2 & 54.0 & 48.9 \\
GPT-4.1 & \underline{89.7} & \textbf{85.0} & \underline{79.8} & \underline{82.0} & \underline{89.2} & 31.1 & \underline{57.9} & 15.4 & \underline{46.0} & 13.1 & \underline{70.0} & \underline{65.9} \\
Gemini-2.0-Flash & \textbf{95.9} & \underline{84.5} & 77.2 & \textbf{90.0} & \textbf{91.1} & \underline{33.9} & \textbf{63.0} & \textbf{36.3} & \textbf{49.9} & \underline{24.3} & \textbf{78.5} & \textbf{70.9} \\
\bottomrule
\end{tabular}
\caption{Scores (\%) of different LLMs on the \textbf{Kazakh} tasks of MiLiC-Eval. 
\textbf{Vocab. Test} refers to Vocabulary Understanding. 
\textbf{TC Sent.} refers to topic classification (Sentence).
\textbf{TC Pass.} refers to topic classification (Passage).
\textbf{Read. Comp.} refers to Reading Comprehension.
\textbf{Resp. Sel.} refers to Response Selection.
\textbf{Title Gen.} refers to Title Generation.
\textbf{MT-A} refers to Machine Translation (Article).
xx2en denotes translation from the minority languages to English.
en2xx denotes translation from English to the minority languages.
\textbf{MT-D} refers to Machine Translation (Dialogue).
\textbf{Math} refers to Math Reasoning.
When calculating the average, the score for the two translation tasks is the mean of xx2en and en2xx. The highest scores are indicated in \textbf{bold}, while the second-highest scores are marked with \underline{underline}.}
\label{tab:main_experiment_kk}
\end{table*}

\begin{table*}[t]
\small
\centering
\setlength\tabcolsep{5pt}
\begin{tabular}{l|ccccccccccc|c}
\toprule
\textbf{Model} &  \textbf{{Vocab.}} & \textbf{{TC}} & \textbf{{TC}} & \textbf{{Read.}} & \textbf{Resp.} & \textbf{Title} & \textbf{MT-A} & \textbf{MT-A} & \textbf{MT-D} & \textbf{MT-D} & \textbf{Math} & \textbf{Avg.}\\
 & \textbf{Test} & \textbf{Sent.} & \textbf{Pass.} & \textbf{Comp.} & \textbf{Sel.} & \textbf{Gen.} & xx2en & en2xx & xx2en & en2xx &  & \\
\midrule
 Random & 25.0 & 16.2 & 25.0 & 27.0 & 25.0 & - & \ \ - & \ \ - & \ \ - & \ \ - & \ \ - & - \\
 \midrule
Ministral-8B & 25.9 & 18.1 & 43.8 & 39.5 & 33.7 & \ \ 7.9 & 16.0 & \ \ 4.4 & 11.6 & \ \ 5.8 & \ \ 7.0 & 21.7 \\
Aya-Expanse-8B & 28.4 & 21.4 & 33.1 & 30.3 & 30.8 & 10.1 & 15.0 & \ \ 6.6 & 13.0 & \ \ 7.6 & \ \ 9.2 & 20.5 \\
Llama-3.1-8B & 26.7 & 23.1 & 39.8 & 31.5 & 25.6 & \ \ 8.7 & 18.4 & \ \ 5.5 & 14.9 & \ \ 6.7 & \ \ 8.7 & 20.8 \\
Qwen-2.5-7B & 31.0 & 30.1 & 37.8 & 44.0 & 40.4 & \ \ 9.2 & 18.1 & \ \ 4.8 & 14.1 & \ \ 4.0 & 11.2 & 24.9 \\
Qwen-3-8B & 29.3 & 28.2 & \underline{53.8} & \underline{46.5} & 43.4 & 12.2 & 18.4 & \ \ 4.8 & 14.7 & \ \ 4.8 & \underline{23.5} & \underline{28.7} \\
Qwen-3-14B & 29.7 & 21.5 & 44.4 & 45.0 & 42.8 & 12.3 & 18.4 & \underline{\ \ 9.6} & 15.0 & \underline{10.7} & 22.0 & 27.2 \\
Gemma-2-9B & 30.3 & 19.9 & 45.3 & 42.8 & 37.0 & \underline{12.5} & 17.4 & \ \ 5.1 & 15.5 & \ \ 6.7 & 14.7 & 25.0 \\
Gemma-3-12B & 31.2 & 23.6 & 42.1 & 41.0 & 41.8 & 10.2 & 16.5 & \ \ 7.7 & 11.8 & \ \ 9.0 & 13.0 & 25.1 \\
\midrule
TowerInstruct-7B & 25.9 & 15.0 & \ \ 3.6 & \ \ 0.5 & \ \ 0.0 & \ \ 0.0 & \ \ 9.3 & \ \ 3.9 & \ \ 1.0 & \ \ 3.0 & \ \ 5.0 & \ \ 6.5 \\
EMMA-500-7B & 25.4 & 21.1 & \ \ 1.0 & \ \ 0.0 & \ \ 0.0 & \ \ 0.0 & \ \ 8.6 & \ \ 4.4 & \ \ 7.5 & \ \ 5.5 & \ \ 8.0 & \ \ 7.6 \\
Bayling-2-8B & 23.3 & 19.4 & 35.9 & 28.0 & 29.2 & \ \ 0.9 & 13.9 & \ \ 1.5 & \ \ 9.7 & \ \ 5.5 & \ \ 8.0 & 17.8 \\
LLaMAX-3-8B & 25.9 & \underline{30.8} & 25.2 & 35.0 & 29.5 & \ \ 0.3 & 16.6 & \ \ 3.0 & 12.6 & \ \ 3.1 & \ \ 4.5 & 18.8 \\
\midrule
GPT-4o-mini & 30.2 & 18.5 & 35.5 & 36.5 & 32.9 & \ \ 6.4 & 19.6 & \ \ 0.8 & 16.6 & \ \ 0.8 & \ \ 7.0 & 20.7 \\
GPT-4.1 & \underline{34.1} & 21.1 & 46.2 & 45.0 & \underline{43.5} & \ \ 9.2 & \underline{20.1} & \ \ 0.6 & \underline{16.6} & \ \ 7.2 & 23.5 & 27.2 \\
Gemini-2.0-Flash & \textbf{81.2} & \textbf{82.4} & \textbf{93.1} & \textbf{84.5} & \textbf{86.5} & \textbf{21.8} & \textbf{52.7} & \textbf{12.1} & \textbf{48.1} & \textbf{20.1} & \textbf{85.5} & \textbf{66.8} \\
\bottomrule
\end{tabular}
\caption{Scores (\%) of different LLMs on the \textbf{Mongolian} tasks of MiLiC-Eval. 
\textbf{Vocab. Test} refers to Vocabulary Understanding. 
\textbf{TC Sent.} refers to topic classification (Sentence).
\textbf{TC Pass.} refers to topic classification (Passage).
\textbf{Read. Comp.} refers to Reading Comprehension.
\textbf{Resp. Sel.} refers to Response Selection.
\textbf{Title Gen.} refers to Title Generation.
\textbf{MT-A} refers to Machine Translation (Article).
xx2en denotes translation from the minority languages to English.
en2xx denotes translation from English to the minority languages.
\textbf{MT-D} refers to Machine Translation (Dialogue).
\textbf{Math} refers to Math Reasoning.
When calculating the average, the score for the two translation tasks is the mean of xx2en and en2xx. The highest scores are indicated in \textbf{bold}, while the second-highest scores are marked with \underline{underline}.}
\label{tab:main_experiment_mn}
\end{table*}

\begin{table*}[t]
\small
\centering
\begin{tabular}{l|cccccccc}
\toprule
\textbf{Model} &  \textbf{{TC}} &  \textbf{{Read.}} & \textbf{Resp.} &  \textbf{MT-A} & \textbf{MT-A} & \textbf{MT-D} & \textbf{MT-D} & \textbf{Math} \\
 & \textbf{Sent.}  & \textbf{Comp.} & \textbf{Sel.} &  xx2en & en2xx & xx2en & en2xx & \\
\midrule
\multicolumn{9}{c}{\textit{Tibetan}}\\
\midrule
Qwen-2.5-7B & 30.6  & {48.3} & {39.6} & 18.8 & 10.8 & 14.2 & 13.2 & 13.3 \\
Llama-3.1-8B &  {63.5} & 44.8 & 37.4 &  {27.5} & \ \ 9.3 & {28.3} & 12.8 & {29.7}  \\
Gemma-2-9B &  {61.8} & {58.2} & {53.0}  & {25.2} & {15.6} & {27.8} & {19.6} & {32.7} \\
\midrule
\multicolumn{9}{c}{\textit{Uyghur}}\\
\midrule
Qwen-2.5-7B &  70.6 & 57.0 & 47.8 & 30.8 & 12.2 & 26.4 & 11.7 & 38.0  \\
Llama-3.1-8B &  {76.9}  & {59.2} & 46.0 & {41.3} & 17.6 & 36.2 & 15.9 & {48.0} \\
Gemma-2-9B &  {75.9} & 58.5 & 59.5  & 37.0 & 12.1 & 31.7 & 11.1 & 46.8  \\
\midrule
\multicolumn{9}{c}{\textit{Kazakh}}\\
\midrule
Qwen-2.5-7B &  56.5 & {51.7} & 40.1  & 25.6 & \ \ 1.0 & 21.0 & \ \ 1.1 & 21.3 \\
Llama-3.1-8B &  55.5  & 45.7 & 32.9  & 24.3 & {11.0} & 18.9 & {11.0} & 24.0  \\
Gemma-2-9B &  57.9  & 51.7 & {43.0}  & 23.4 & \ \ 7.8 & 19.9 & \ \ 8.7 & {29.3}   \\
\midrule
\multicolumn{9}{c}{\textit{Mongolian}}\\
\midrule
Qwen-2.5-7B & {30.1} &{44.0} & {40.4}  & 18.1 & \ \ 4.8 & 14.1 & \ \ 4.0 & {11.2}  \\
Llama-3.1-8B &  23.1  & 31.5 & 25.6 & {18.4} & {\ \ 5.5} & 14.9 & \ \ 6.7 & \ \ 8.7 \\
Gemma-2-9B &  19.9  & {42.8} & {37.0} & 17.4 & \ \ 5.1 & {15.5} & {\ \ 6.7} & {14.7} \\
\midrule
\multicolumn{9}{c}{\textit{Chinese}}\\
\midrule
Qwen-2.5-7B &  87.1 & 92.5 & 90.6  & 54.0 & 25.3 & 64.2 & 23.4 & 76.3 \\
Llama-3.1-8B & 86.3  & 90.2 & 83.8  & 49.5 & 25.0 & 62.6 & 24.0 & 69.5  \\
Gemma-2-9B  & 88.5  & 94.2 & 92.2  & 56.8 & 31.2 & 69.4 & 25.9 & 64.3  \\
\midrule
\multicolumn{9}{c}{\textit{English}}\\
\midrule
Qwen-2.5-7B & 86.6 & 90.5 & 86.3  & - & - & - & - & 71.2  \\
Llama-3.1-8B & 86.5  & 87.8 & 85.1 & - & - & - & - & 75.0 \\
Gemma-2-9B & 88.0  & 93.7 & 92.4  & - & - & - & - & 65.7 \\
\bottomrule
\end{tabular}
\caption{Scores (\%) of three LLMs on the tasks with cross-lingual parallism in MiLiC-Eval. 
\textbf{TC Sent.} refers to topic classification (Sentence).
\textbf{Read. Comp.} refers to Reading Comprehension.
\textbf{Resp. Sel.} refers to Response Selection.
\textbf{MT-A} refers to Machine Translation (Article).
xx2en denotes translation from the minority languages to English.
en2xx denotes translation from English to the minority languages.
\textbf{MT-D} refers to Machine Translation (Dialogue).
\textbf{Math} refers to Math Reasoning.
.}
\label{tab:comparison_with_hrl}
\end{table*}

\end{document}